\documentclass[10pt,twocolumn,letterpaper]{article}

\usepackage{wacv}
\usepackage[accsupp]{axessibility} 
\usepackage{times}
\usepackage{epsfig}
\usepackage{graphicx}
\usepackage{amsmath}
\usepackage{amssymb}
\usepackage{booktabs}
\usepackage{algorithm}
\usepackage{multirow}
\usepackage[noend]{algpseudocode}
\usepackage[accsupp]{axessibility}

\wacvfinalcopy

\ifwacvfinal
\usepackage[breaklinks=true,colorlinks,bookmarks=false]{hyperref}
\else
\usepackage[pagebackref=true,breaklinks=true,colorlinks,bookmarks=false]{hyperref}
\fi

\begin{document}

\title{Federated Learning for Commercial Image Sources}

\author{Shreyansh Jain\\
IIIT Delhi\\
New Delhi, India\\
{\tt\small shreyansh21089@iiitd.ac.in}
\and
Koteswar Rao Jerripothula\\
IIIT Delhi\\
New Delhi, India\\
{\tt\small koteswar@iiitd.ac.in}
}

\maketitle
\thispagestyle{empty}

\begin{abstract}
Federated Learning is a collaborative machine learning paradigm that enables multiple clients to learn a global model without exposing their data to each other. Consequently, it provides a secure learning platform with privacy-preserving capabilities. This paper introduces a new dataset containing 23,326 images collected from eight different commercial sources and classified into 31 categories, similar to the Office-31 dataset. To the best of our knowledge, this is the first image classification dataset specifically designed for Federated Learning. We also propose two new Federated Learning algorithms, namely Fed-Cyclic and Fed-Star. In Fed-Cyclic, a client receives weights from its previous client, updates them through local training, and passes them to the next client, thus forming a cyclic topology. In Fed-Star, a client receives weights from all other clients, updates its local weights through pre-aggregation (to address statistical heterogeneity) and local training, and sends its updated local weights to all other clients, thus forming a star-like topology. Our experiments reveal that both algorithms perform better than existing baselines on our newly introduced dataset.
\end{abstract}

\section{Introduction}

Federated Learning\footnote{\textit{Published in the IEEE/CVF WACV 2023 Proceedings with DOI: \href{https://doi.org/10.1109/WACV56688.2023.00647}{10.1109/WACV56688.2023.00647}.  
Please cite this work as \cite{10030788}. \\
© 2023 IEEE. Personal use of this material is permitted. Permission from IEEE must be
obtained for all other uses, in any current or future media, including
reprinting/republishing this material for advertising or promotional purposes, creating new
collective works, for resale or redistribution to servers or lists, or reuse of any copyrighted
component of this work in other works}} (FL) is a distributed learning paradigm that can learn a global model from decentralized data without having to exchange sensitive data across the clients ~\cite{li2020federated}~\cite{mcmahan2017communication}. 

Traditional Machine Learning requires users to upload their data to the centralized server for the learning and inference task. The end-user has no power and control over how the data is used~\cite{drainakis2020federated}. Moreover, uploading the data to a central server incurs severe costs. Maintaining such a vast volume of data and communicating the learning parameters back to the user is costly. To overcome the privacy challenges and issue of maintaining a large amount of data in the centralized setting, Google~\cite{konevcny2016federated} proposed the Federated Learning paradigm, which could handle these issues well.

The Federated Learning framework addresses sensitive data privacy and data access issues~\cite{truex2019hybrid}. Federated Learning models are trained via model aggregation rather than data aggregation. It requires model to be trained locally on the data owner's machine or the local edge devices, and only the model parameters are shared. 
Federated Learning has found successful applications in the IoT, healthcare, finance, etc~\cite{li2020review}~\cite{nguyen2021federated}.
The traditional Federated Learning optimization methods involve local client training on the local datasets for a fixed number of epochs using an SGD optimizer. The local clients then upload the model weights to the central server, where the weights are averaged to form a global model whose parameters are shared with the local client. This method is known as FedAvg~\cite{mcmahan2017communication}, which facilitates the local client to learn features from different clients while preserving privacy. However, FedAvg may have convergence issues in case the clients exhibit statistical heterogeneity, which may lead to non-convergence of the model~\cite{li2019convergence}~\cite{karimireddy2019scaffold}~\cite{hsu2019measuring}. Thus, a simple FedAvg algorithm may not be helpful when dealing with device-level heterogeneity.

In this work, we aim to understand the real-world scenario where different commercial image sources can collaborate in a Federated setting to perform the image classification task with privacy preservation.

Most previous research works applied the Federated Learning algorithm on a single dataset distributed among the clients in an IID or non-IID manner. This is not close to a real-world scenario where different clients may have different data distribution due to domain shift (statistical heterogeneity) among them ~\cite{quinonero2008dataset} ~\cite{moreno2012unifying}. Another challenge in Federated Learning is the convergence issue when the data distribution is different among the clients, which may increase the communication cost between the clients and the central server and leads to suboptimal model performance.

Motivated by the above challenges, we propose our dataset in which each client's dataset is sampled from different commercial image sources to simulate the real-world scenario where each client exhibits a domain shift. This is because each commercial image source has its own unique image set, causing domain shift amongst clients. The dataset is inspired by the Office-31 dataset ~\cite{saenko2010adapting}. We also propose two novel algorithms, namely Fed-Cyclic and Fed-Star. Fed-Cyclic is a simple algorithm in which, a client gets weights from the previous client, trains the model locally and passes the weights to the next client in a cyclic fashion. In this way, a global model is simply being passed from one client to another in a cyclic manner. The global server need not involve here. Even if it is required that we want to involve it to preserve anonimity, the global server does not have to perform any computation and can be used only for passing the parameters of one client to another. Fed-Star requires each client to receive weights from all the other clients during pre-aggregation in a star-like manner after the local training of each client on its train set. While pre-aggregating, every client prioritizes learning the outlier features present in different clients while retaining the common features to train a more robust model impervious to statistical heterogeneity among the client's data distribution, followed by aggregation via a global server after a fixed number of periods. Experiments show that our algorithms have better converegence and accuracy. 

To summarize, our contributions are three-fold:
\begin{itemize}
\item We propose the an image classification dataset specifically designed
for Federated Learning, which is close to a real-world scenario where each client has a unique dataset demonstrating domain shift.
\item We propose Fed-Cyclic, a simple algorithm, which is communicationally efficient and attains higher accuracy than baselines.

\item We propose the Fed-Star algorithm, which trains a model that prioritizes learning of generalized and outlier features to create a model personalized to each client's heterogeneous dataset distribution and attains faster convergence than the baselines with higher accuracy.
\end{itemize}
\section{Related Works}
Many distributed optimization algorithms have been developed to process and draw inferences from the data uploaded~\cite{zhang2015deep},~\cite{richtarik2016distributed},~\cite{dekel2012optimal}~\cite{boyd2011distributed}~\cite{amiri2020survey}. However, such distributed method requires uploading of data to the central server, which incurs the considerable cost of maintaining data centrally, and processing it requires a lot of power~\cite{papernot2016towards}~\cite{de2020overview}. Also, the privacy issue persists as the user has no control over how personal data is used and shared.

The first application of the Federating Learning algorithm is FedAvg, proposed by Mcmahan \etal~\cite{mcmahan2017communication}.
FedAvg performs reasonably well when the data distribution is IID among the clients and shows faster convergence of the global model. The issue arises in real-world scenarios when the data follow the non-IID distribution as proposed by Zhao \etal~\cite{zhao2018federated}.

\textbf{Federated Learning with data heterogeneity: } The vanilla FedAvg faces convergence issues when there is data heterogeneity among the clients. To tackle this challenge, different methods have been proposed. FedProx~\cite{li2020federated} adds a proximal term by calculating the square distance between the server and client with local loss to optimize the global model better.
FedNova ~\cite{wang2020tackling}  proposes normalized averaging to eliminate objective inconsistency with heterogeneous Federated optimization. FedMax, as proposed by Chen \etal~\cite{chen2020fedmax}, aims to mitigate activation-divergence by making activation vectors of the same classes across different devices similar. FedOpt proposed by Reddi \etal~\cite{reddi2020adaptive} applies different optimizers to the server, like Adam, Yogi, and AdaGrad.
VRL-SGD~\cite{liang2019variance} incorporates variance-reduction into local SGD and attains higher accuracy while decreasing the communication cost. FedCluster~\cite{chen2020fedcluster} proposes grouping local devices into different clusters so that each cluster can implement any Federated algorithm. RingFed, proposed by Yang \etal~\cite{yang2021ringfed}, minimizes the communication cost by pre-aggregating the parameters among the local clients before uploading the parameter to the central server.
SCAFFOLD ~\cite{karimireddy2020scaffold} uses variance-reduction to mitigate client drift.
Chen \etal~\cite{chen2021fedsvrg} proposes FedSVRG that uses stochastic variance reduced gradient-based method to reduce the communication cost between clients and servers while maintaining accuracy. Jeong \etal~\cite{jeong2018communication} proposes Federated augmentation (FAug), which involves the local client jointly training the generative model to augment their local dataset and generate the IID dataset. In this paper, we aim to train models robust to data heterogeneity with faster convergence and lower communication costs via our proposed algorithms. One of our proposed method, Fed-Star is similar to RingFed~\cite{yang2021ringfed}. RingFed involves simple pre-aggregation of weights between adjacent clients, whereas our method involves pre-aggregation of weights between all the local clients using the accuracy metric.

\textbf{Personalized Federated Learning } FedAvg also suffers from creating a generalized global model as the parameters are averaged, which gives poor representation to a client with heterogeneous data.
Personalized Federated Learning involves training the global model using any Federated vanilla algorithm followed by personalizing the model for each client via locally training the model on each client ~\cite{kairouz2021advances}~\cite{mansour2020three}~\cite{zhao2018federated}~\cite{drainakis2020federated}. Data heterogeneity among clients is the reason for personalized Federated Learning. Data augmentation is explored to account for local data heterogeneity and involves local clients to jointly generate IID data distribution~\cite{duan2020self}~\cite{wu2020fedhome}.
Wang \etal~\cite{wang2020optimizing} proposes FAVOR and selects a subset of clients at each round to mitigate the bias introduced by non-IID data. Chai \etal~\cite{chai2020tifl} proposes the TiFL method to cluster the clients in different tiers and train the clients belonging to the same tier together for faster training.  Sattler \etal~\cite{sattler2020clustered} proposes a hierarchical clustering-based approach based on the cosine similarity of the client gradient to segment similar clients in similar clustering for training to train the local clients properly. Xie \etal ~\cite{xie2020multi} proposes Expectation Maximization to derive optimal matching between the local client and the global model for personalized training. Deng \etal~\cite{deng2020adaptive} proposes APFL algorithm to find optimal client-server pair for personalized learning.There are different clustering-based approaches as explored by different authors ~\cite{briggs2020federated}~\cite{ghosh2020efficient}~\cite{huang2019patient}~\cite{duan2021fedgroup}. Tan \etal~\cite{tan2022towards} provides deeper analysis of personalized Federated framework.
In our work, we aim to personalize the model at the global level by proposing an algorithm that better captures the outlier features of the client while retaining the generalized features.

\textbf{Federated Image classification datasets}
Most Federated Learning algorithms are simulated on datasets belonging to a single domain with an artificial partition among the clients or use existing public datasets. The dataset distribution may differ for different clients in the real-world scenario as the clients exhibit domain shift. The first work proposing the real-world image dataset ~\cite{luo2019real} contains more than 900 images belonging to 7 different object categories captured via street cameras and annotated with detailed boxes. The image dataset has applications in object detection. In our work, we propose the first real-world image dataset for the image classification task to better understand the performance of the Federated algorithm in a real-world setting.

\section{Proposed Methods}
\subsection{Objective}

In Federated Learning (FL), different clients (say $K$ clients) collaborate to learn a global model without having to share their data. Let the weights of such a model be $w$, and let the loss value of the model for sample $(x_i,y_i)$ be $\mathcal{L}(x_i,y_i;w)$. The objective now is to find optimal $w$ such that the following objective is achieved:
\begin{equation}
    \min \frac{1}{|D|}\sum\limits_{i=1}^{|D|}{\mathcal{L}(x_i,y_i;w)}
\end{equation}
where $D$ denotes the union of all the data owned by different clients, as shown below:
\begin{equation}
D=\bigcup\limits_{k=1}^{K}{D_k}
\end{equation}
where $D_k$ denotes the data owned by the $k^{th}$ client. Given this, we can rewrite our objective function as follows:
\begin{equation}
    \min \frac{1}{|D|}\sum\limits_{k=1}^{K}{\sum\limits_{i=1}^{|D_k|}\mathcal{L}(x_i,y_i;w)}
\end{equation}

If we represent the average local loss $\mathbf{L}_k$ of $k^{th}$ client using 

\begin{equation}
\mathbf{L}_k=\frac{1}{|D_k|}\sum\limits_{i=1}^{|D_k|}\mathcal{L}(x_i,y_i;w),
\end{equation}
we can reformulate the objective as follows:
\begin{equation}
    \min \sum\limits_{k=1}^{K}{\frac{|D_k|}{|D|}\mathbf{L}_k}
\end{equation}
which suggests that our objective is to minimize the weighted sum of local losses incurred by our clients, and the weights are proportional to the number of data samples clients have with them. This objective function is similar to same as ~\cite{mcmahan2017communication}. We discussed it to make our paper self-contained. 

This formulation motivated FedAvg ~\cite{mcmahan2017communication} to aggregate the local model weights in the weighted averaging manner to obtain $w$ as shown below:

\begin{equation}
w=\sum\limits_{k=1}^K{\frac{|D_k|}{|D|}w_k}
\end{equation}

\begin{figure}[!htb]
  \centering
  \includegraphics[width=1.0\linewidth]{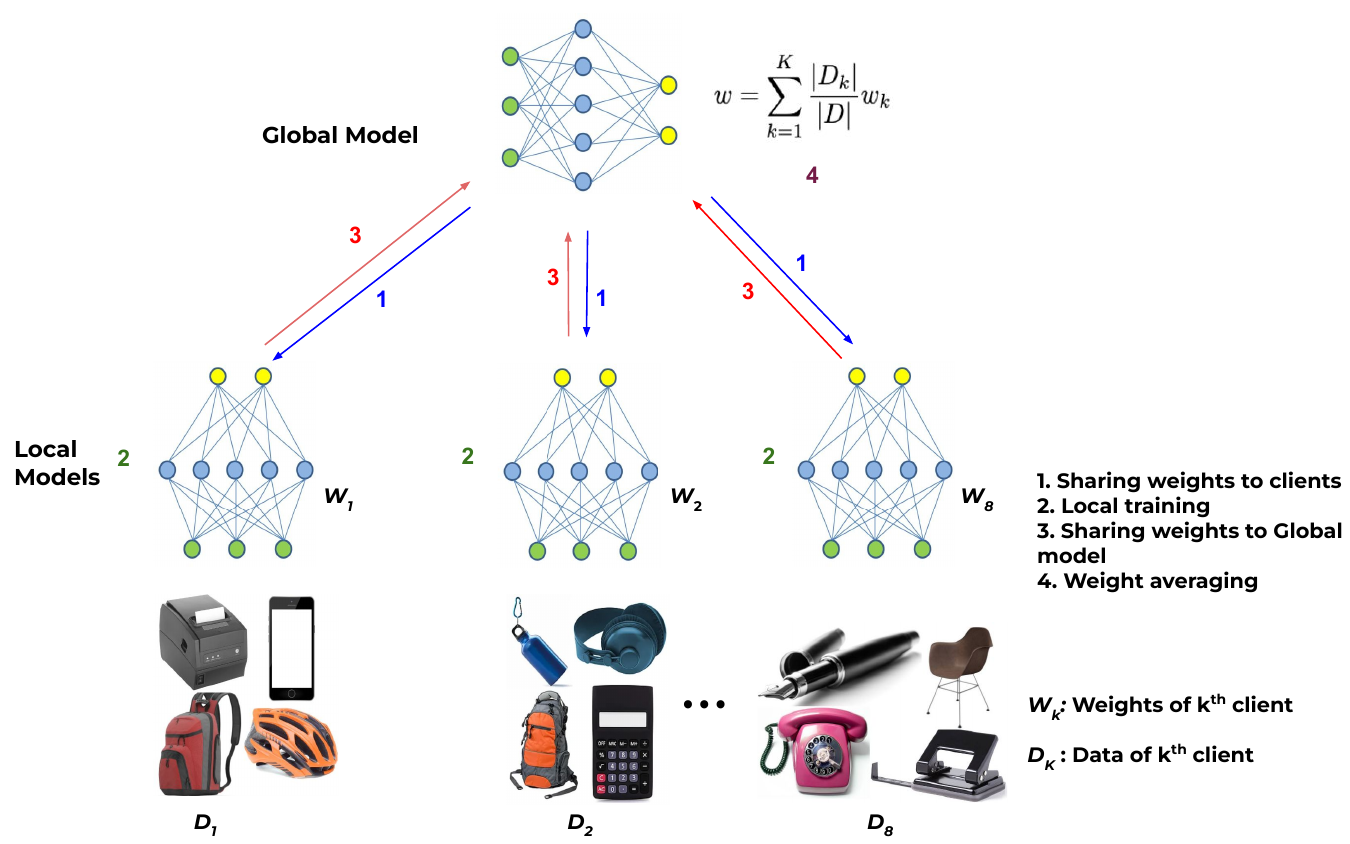}
  \caption{FedAvg algorithm ~\cite{mcmahan2017communication}}
  \label{fig:fed_avg}
\end{figure}

This aggregation happens iteratively, where, in a given iteration, the central server sends the global model to the local clients, where local models are updated and then sent back to the global server for aggregation, as shown in Figure~\ref{fig:fed_avg}. This happens until the global model converges. 

However, it can take several communication rounds for the FedAvg algorithm to converge, especially when there is statistical heterogeneity in clients' datasets. As a result, the accuracy drops too~\cite{han2015deep}. Also, FedAvg creates a generalized model by averaging the parameters from the local clients, forcing the local model with statistical heterogeneity to learn a generalized representation that may differ from its data distribution, leading to poorly trained local clients. In that case, local clients do not find the global model satisfactory. 

Considering these limitations of FedAvg, we propose Fed-Cyclic and Fed-Star algorithms, which cater to the statistical heterogeneity of data across the clients and ensure the satisfactory local performance of global models despite that. 

\subsection{Fed-Cyclic}
\begin{figure}[H]
  \centering
  \includegraphics[width=1.0\linewidth]{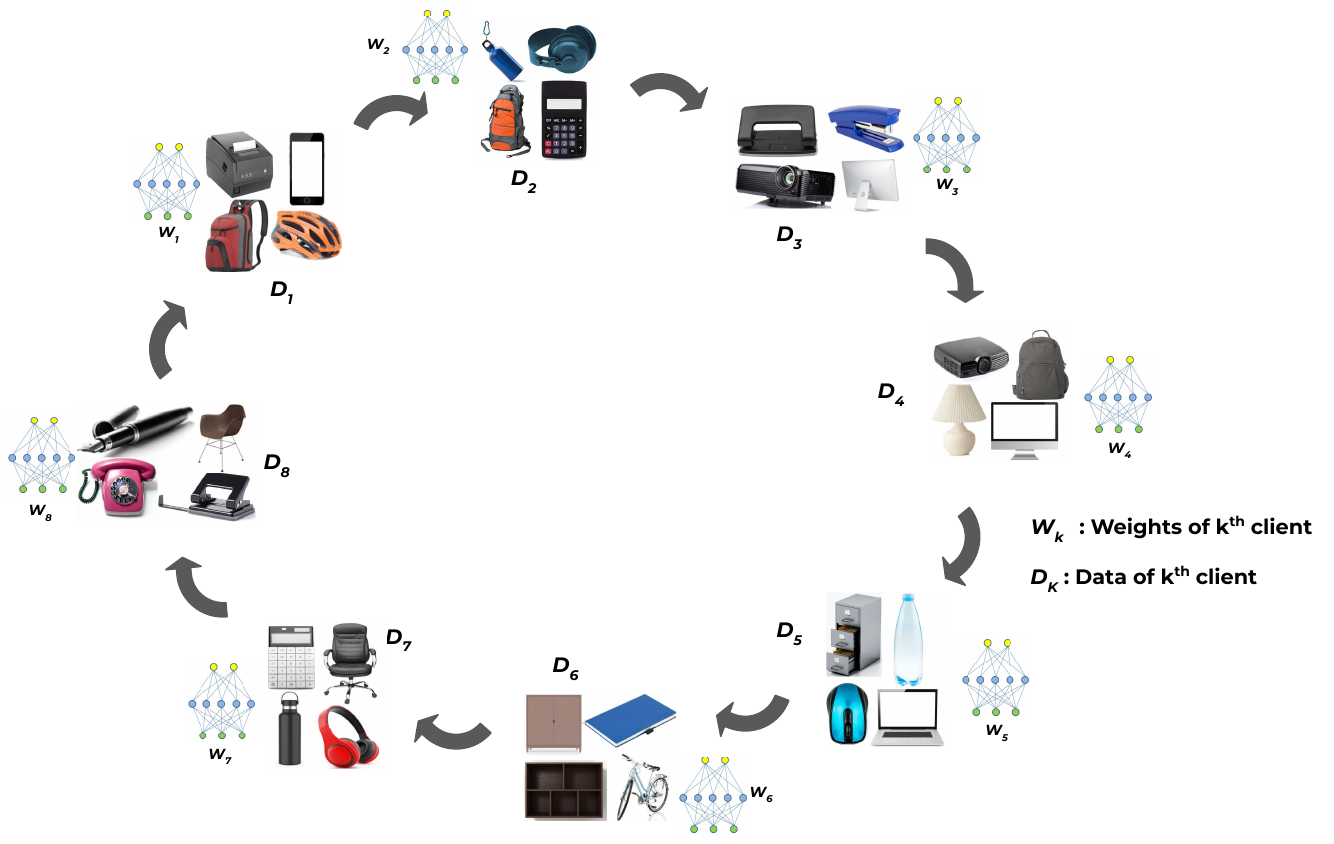}
  \caption{Proposed model using Fed-Cyclic showing passing of weights cyclically by the clients after local training.}
  \label{fig:fed_cyclic}
\end{figure}

We propose the Fed-Cyclic algorithm to overcome the challenges faced by the FedAvg algorithm, which suffers from a communication bottleneck due to a large number of edge devices uploading the parameters to the central server, which causes congestion in the network. The model visualization is shown in Figure \ref{fig:fed_cyclic}

    \begin{algorithm}
    \caption{Proposed Fed-Cyclic Algorithm}\label{FedCyclic}
    \hspace*{\algorithmicindent} \textbf{Input} \textsf{ Initial weights $w_{init}$, number of global rounds $R$, number of local epochs $E$, learning rate $\eta$, $K$ clients indexed by $k$ (with local data $D_k$ and local weights $w_k$) and local minibatch size $b$.} \\ 
    \hspace*{\algorithmicindent} \textbf{Output} Global weights $w^R$ (after $R$ rounds)
    
    \hspace*{\algorithmicindent} \textbf{Algorithm:}
    \begin{algorithmic}
    
    \State  Initialize  $w^0\gets w_{init}$ // Global weights initialized
     \State $w^0_1\gets w^0$
    
    \For{\textit{$r$=0 to $R-1$}}
        \For{\textit{$k$=1  to  $K-1$}}
          
        \State \textit{$w^{r+1}_{k}$} $\gets$ $ClientUpdate(w^r_k,k)$ using SGD

        \State \textit{$w^{r}_{k+1}$} $\gets$ \textit{$w^{r+1}_{k}$}
      \EndFor
      \State \textit{$w^{r+1}_{K}$} $\gets$ $ClientUpdate(w^r_K,K)$ using SGD

        \State \textit{$w^{r+1}_{1}$} $\gets$ \textit{$w^{r+1}_{K}$}
         \State \textit{$w^{r+1}$} $\gets$ \textit{$w^{r+1}_{K}$}
      
       \EndFor
       \Function{$ClientUpdate$}{$w,k$}       
       \State $B$ $\gets$ (split $D_k$ into batches of size $b$)
      \For{$e$=1 to $E$}
      \For{$d$ $\in$ $B$}
      $w$ $\gets$ $w$ - $\eta$$\nabla$g(w;$d$)
      \EndFor
      \EndFor
       \State return $w$ 
    \EndFunction
    \end{algorithmic}
    \end{algorithm}
    
In the Fed-Cyclic algorithm, we use a global model to initialize the weight of one of the clients in the network, followed by training the client's local model for $E$ local epochs. The optimizer used is SGD at the local client. The updated weights are then used to initialize the weight of the next client in the network, as shown in equation \eqref{eq:5}, and the process continues until all the clients are trained cyclically in this manner, constituting one training round. In our Fed-Cyclic algorithm, the clients can either directly pass the weights to the next client or involve the global server to do so to preserve the anonymity of the last client. After the end of each round, the global weights $w$ get updated.


It is a communication-efficient algorithm since the clients can directly pass the weights to the next client without involving the global server. Even if the global server is involved in passing the weights from one client to another, no processing is done, and only parameters from a single client are passed at a time. We explain it in Algorithm \ref{FedCyclic} in greater detail. The most important step is the following:

\begin{align}
w^{r}_{k+1} \gets w^{r+1}_{k} 
\label{eq:5}
\end{align}
where we use $k^{th}$ client to initialize $(k+1)^{th}$ client. 

The algorithm is robust to statistical heterogeneity as every client gets an opportunity to train the global model on the local data. Moreover, we can take the view that the global model is being periodically trained on different portions of the dataset ($D$), as if they are mini-batches ($D_k$). Hence, this algorithm is somewhat analogous to a typical deep learning approach from the point of view of the global model. As a result, convergence also gets ensured, unlike FedAvg, where we expect convergence for simple aggregation of weights.
\subsection{Fed-Star}


    \begin{algorithm}
    \caption{Proposed Fed-Star Algorithm}\label{FedStar}

        \hspace*{\algorithmicindent} \textbf{Input:}  Initial weights $w_{init}$, number of global rounds $R$, number of local epochs $E$, learning rate $\eta$, $K$ clients indexed by $k$ (with local data $D_k$ and local weights $w_k$), local minibatch size $b$, number of period $P$, and weight matrix $M$ of dim $K*K$. \\ 
    \hspace*{\algorithmicindent} \textbf{Output:} Global weights $w^R$ (after $R$ rounds)

    \hspace*{\algorithmicindent} \textbf{Algorithm:}
         \begin{algorithmic}
    \State Initialize  $w^0\gets w_{init}$ // Global weights initialized

    \For{$r$=0 to $R-1$}
    \State $w^{r,0}_k\gets w^r, \forall k\in\{1,\cdots,K\}$
    \For{$p \in \{0,\cdots,P-1\}$}
     
    \For{$k\in \{1,\cdots,K\}$ \textbf{parallely}}

     \State $w^{r,p+1}_{k}$ $\gets$ $ClientUpdate$ $(w^{r,p}_k,k)$

        \For{$j\in \{1,\cdots,K\}$ }
       \State transfer $w^{r,p+1}_j$ to $k^{th}$ client
       \State $M(k,j) = 1-Acc(w^{r,p+1}_j,D_k)/100$
       \EndFor
       \State $w_{k}^{r,p+1}=\frac{\sum\limits_{j=1}^{K} M(k,j)*w_{j}^{r,p+1}}{\sum\limits_{j=1}^{K} M(k,j)}$
       \EndFor
       
       \EndFor
       
     \State $w^{r+1}=\frac{1}{K}\sum\limits_{k=1}^K\frac{|D_k|}{|D|}{w^{r,P}_k}$
     
    \EndFor

       \Function{$ClientUpdate$}{$w,k$}
       
       \State $B$ $\gets$ (split $D_k$ into batches of size $b$)
      \For{$e$=1 to $E$}
      \For{$d$ $\in$ $B$}
      $w$ $\gets$ $w$ - $\eta$$\nabla$g(w;$d$)
      \EndFor
      \EndFor
      
    \State return $w$ 
    \EndFunction

    \Function{ $Acc$}{$w,D$} 
    \State return Accuracy of $w$ on train set of $D$
    \EndFunction
    \end{algorithmic}
    \end{algorithm}
    
Although Fed-Cyclic can converge faster than FedAvg, it is a very simple algorithm, similar to FedAvg. Also, it lacks aggregation of any kind. Here, we propose the Fed-Star algorithm, where we address these limitations.         

In Fed-Star, for some time, the local models are trained locally for some epochs in a parallel manner. Once the given epochs are complete, we perform pre-aggregation of weights locally at each client by sharing their models with each other. Each client gets models from every other client, and they are evaluated on the local training set of the given client. The accuracy obtained helps us determine how much weightage should be given to the models of each client during pre-aggregation. These pre-aggregated weights are now used to reinitialize the local model for training. These steps are iteratively carried out, and these iterations are called periods. This interaction for model sharing is analogous to star network topology, where every client interacts with every other client in a network. After a certain number of periods $P$, the local weights are aggregated on the central servers, so a round has $P$ periods in it, where local models are being shared with each other, they are getting pre-aggregated at each client for initialization to train the model in the next period.      

In any period $p$ of round $r$, a weightage matrix $M$ gets developed, which is computed as follows:
\\
\begin{equation}
   M(k,j) = 1-Acc(w^{r,p+1}_j,D_k)/100 
\end{equation}
where $Acc(w^{r,p+1}_j,D_k)$ denotes the training accuracy of $w^{r,p+1}_j$ on training set of dataset $D_k$. It denotes the weightage value for the model coming from $j^{th}$ model while pre-aggregating at $k^{th}$ client.      


Note here that we give more weightage to the client significantly different from the reference client during pre-aggregation because we want each client to learn the outlier features from the clients while retaining the generalized features during pre-aggregation. 


The pre-aggregated weights for $k^{th}$ client are depicted as follows:

\begin{align}
w_{k}^{r,p+1}=\frac{\sum\limits_{j=1}^{K} M(k,j)*w_{j}^{r,p+1}}{\sum\limits_{j=1}^{K} M(k,j)}
\label{eq:6}
\end{align}
where we normalize the weightages with their sum while performing the pre-aggregation of weights. 

\begin{figure}[t]
  \centering
  \includegraphics[width=1.0\linewidth]{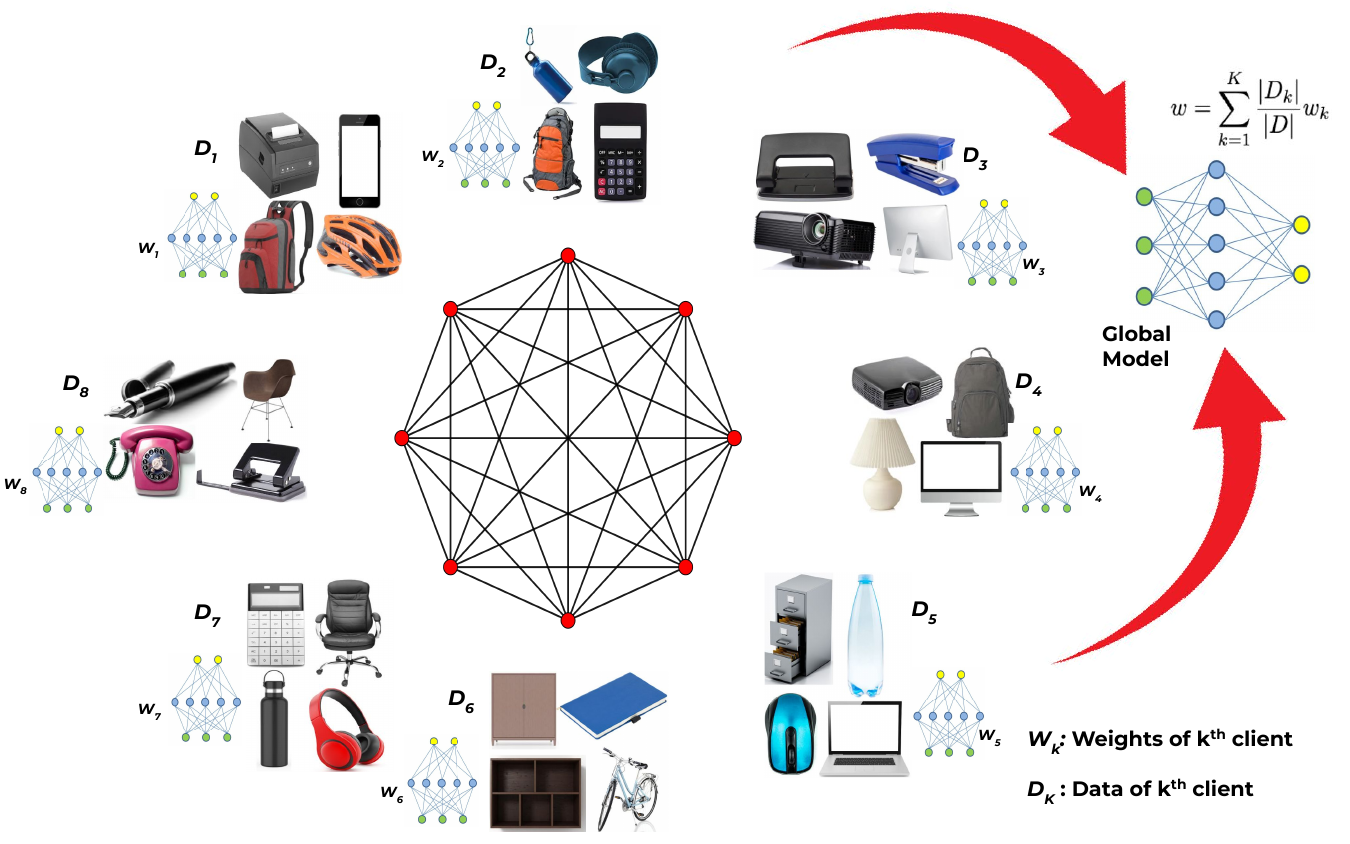}
  \caption{Proposed model using Fed-Star demonstrating pre-aggregation of the parameters in star topology manner among local clients and is given by equation \eqref{eq:6}. This is followed by the transferring of weight to the global model where aggregation of weights is performed.}
  \label{fig:fed_star}
\end{figure}

Thus, each local model tends to learn more from the other client that is significantly different. The Fed-Star algorithm is explained further in Algorithm \ref{FedStar} and can be visualized through Figure \ref{fig:fed_star}. 

This algorithm is communication intensive since each client has to interact with every other client. Still, the total communication overhead is reduced significantly as the pre-aggregation step among clients decreases the reliance on the global server for convergence. Our algorithm attains faster convergence than FedAvg with lesser communication overhead with the global server and higher accuracy. Fed-Star retains outlier features well and helps create a global model that is also personalized to the local clients.



\section{Proposed Dataset}
\begin{figure}[!htb]
  \centering
  \includegraphics[width=0.85\linewidth]{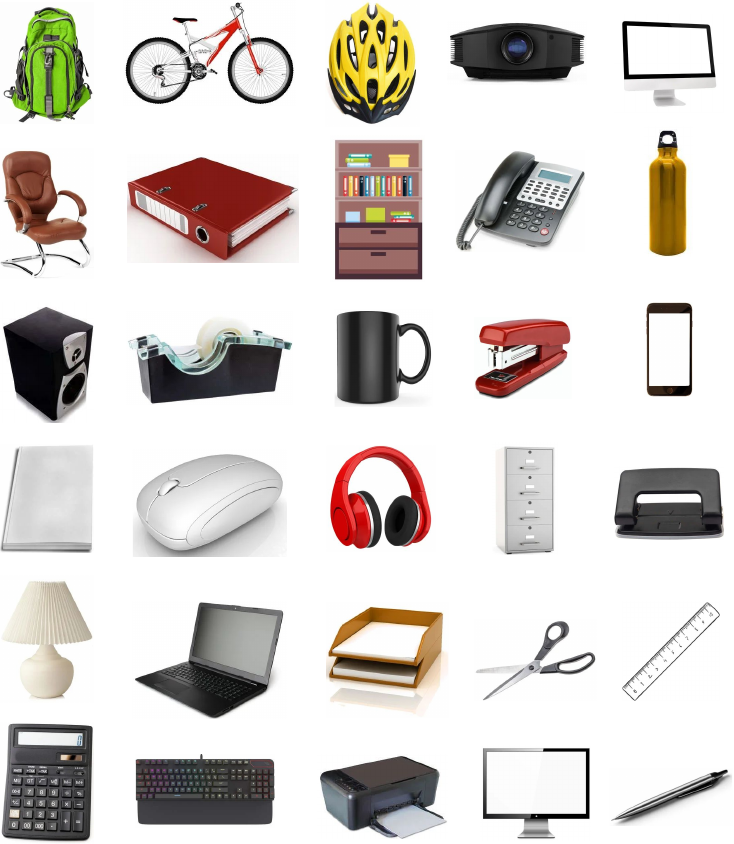}
  \caption{Sample images from our dataset.}
  \label{fig:sample}
\end{figure}

\begin{table}[!htb]
  \begin{center}
    \scalebox{0.85}{{\small{
\begin{tabular}{lrrr}
\toprule
Classes & Mean & Std. dev. & Total Img. \\
\midrule
back\_pack   &  108.50  &   37.98 & 868\\
bike   & 108.62 & 31.73 & 869\\
bike\_helmet   & 94.25 & 26.54 & 754\\
book\_shelf & 77.13 & 25.93 & 617\\
bottle & 89.38 & 11.61  & 715\\
calculator & 111.50 & 31.43 & 892\\
desk\_chair & 125.38 & 27.61 & 1003\\
desk\_lamp  & 106.87 & 24.87 & 855\\
desktop\_computer   &  99.00  &   19.13 & 792 \\
file\_cabinet   &  65.38  &   20.13 & 523 \\
headphone   &  105.63  &   22.36 & 845\\
keyboard   &  76.38  &   35.14 & 611\\
laptop   &  111.38  &   25.39 & 891\\
letter\_tray   &  26.38  &   13.47 & 211 \\
mobile\_phone   &  84.88  &   15.06 & 679 \\
monitor   &  112.38  &   37.35 & 899\\
mouse   &  116.87  &   25.97 & 935\\
mug   &  117.38  &   25.19 & 939\\
notebook   &  98.00  &   23.78 & 784 \\
pen   &  108.50  &   25.20  & 868\\
phone   &  113.63  &   21.80 & 909\\
printer   &  108.25  &   22.64 & 866\\
projector   &   72.63   &   33.86 & 581\\
puncher   &  66.88  &   32.37 & 535\\
ring\_binder   &  65.63 &   31.26 & 525\\
ruler   &  90.13  &   19.50 & 721 \\
scissors   &  127.13  &   51.77 & 1017\\
speaker   &  67.63  &   18.6 & 541 \\
stapler   &  105.25  &   21.75 & 842 \\
tape\_dispenser   &  82.38  &   37.9 & 659 \\
trashcan   &  103.25  &   33.36 & 826\\
\bottomrule
\end{tabular}
}}}
\end{center}
\vspace{0.01cm}
\caption{The mean, standard deviation and the total number of images in the classes across different commercial sources.}
\label{stats1}
\end{table}
We propose a dataset containing 23,326 images which we collected from 8 different image-hosting websites. Each of the 8 sources represents 8 different clients in our Federated learning setting. The average number of images for each source is approximately 2916, divided across 31 categories. This dataset is inspired from the Office-31 dataset~\cite{saenko2010adapting}, which contains common objects in the office settings like keyboard, printer, monitor, laptop, and so forth. Our dataset includes the same categories of images as were in the Office-31 dataset. We took extra care to ensure that only relevant and high-quality images from each source were taken. We removed poor quality, duplicate or irrelevant images by manually curating the dataset. The statistics showing how images are distributed within the classes and across the sources are summarized in Tables~\ref{stats1}~and~\ref{stats2}.
The sample images are shown in Figure~\ref{fig:sample}.

\begin{table}[!htb]
  \begin{center}
    {\small{
\begin{tabular}{lrrr}
\toprule
Source & Mean & Std. dev. & Total Img. \\
\midrule
\href{https://www.123rf.com/}{123rf}   &  94.16  &   26.96 & 2897 \\
\href{https://stock.adobe.com/}{Adobe Stock}   & 101.16 & 27.25 & 3104\\
\href{https://www.alamy.com/}{Alamy}   & 102.80 & 	40.65 & 3155 \\
\href{https://www.canstockphoto.com/}{CanStockPhotos} & 95.06 & 33.24 & 2915\\
\href{https://depositphotos.com/}{Depositphotos} & 101.41 & 38.59 & 3112\\
\href{https://www.gettyimages.in/}{Getty Images}  & 63.06 & 21.99 & 1923\\
\href{https://www.istockphoto.com/}{iStock} & 90.10 & 33.01 & 2761 \\
\href{https://www.shutterstock.com/}{Shutterstock}  & 112.61 & 36.34 & 3459\\
\bottomrule
\end{tabular}
}}
\end{center}
\vspace{0.01cm}
\caption{The mean, standard deviation of images distribution across the classes of the dataset together with total images in each source.}
\label{stats2}
\end{table}
\section{Experiments}
\subsection{Implementation Detail}
In this section, we describe the experiments we performed to evaluate our Federated image classification algorithms. We leverage the pretrained VGG-19~\cite{simonyan2014very} network available from PyTorch pretrained model library ~\cite{paszke2019pytorch} for initialization purpose. We freeze its convolutional layers and replace the rest of the network with three new fully connected layers (of size 1024, 256 and 31) and a softmax layer. In the first two fully connected layers, we use ReLU activation and a dropout rate of 0.5. We have used SGD optimizer with a batch size of 64. We use the 80:20 train-test split of the data at any client. The evaluation metric used is classification accuracy, but we have also evaluated using Macro F1 score and weighted F1 score. The default learning rate is 3e-4 (3x10$^{-4}$). The different iteration parameters used are given in Table~\ref{stats3}.

\begin{table}[ht]
  \begin{center}
    {\small{
\begin{tabular}{lccc}
\toprule
Method & Local Epochs ($E$)  & Period ($P$) & Global Rounds ($R$)  \\
\midrule
FedAvg   &  3  &  - & 250 \\
RingFed  &  3 & 2 & 50 \\
Fed-Cyclic & 3 & - & 150 \\ 	 Fed-Star & 3 & 2& 50\\
\bottomrule
\end{tabular}
}}
\end{center}
\caption{Iteration Parameters}
\label{stats3}
\vspace{3 mm}
  \begin{center}
    {\small{
\begin{tabular}{cc}
\toprule 
 $\gamma$ & Accuracy\\
 \midrule
 0.2 & 89.22\% \\ 
 0.5 & 88.96\% \\
 0.8  & \textbf{89.65\%}  \\
 1.0  & 89.39\%  \\ 
 \bottomrule
\end{tabular}
}}
\end{center}
\caption{$\gamma$ vs accuracy for RingFed}
\label{mul_ring}
\vspace{3 mm}
  \begin{center}
    {\small{
\begin{tabular}{lcccc}
\toprule 
 Method & Accuracy & Weighted F1 & Macro F1 \\
 \midrule
 FedAvg~\cite{mcmahan2017communication} & 89.11\% & 88.98\% & 88.47\% \\ 
 RingFed~\cite{yang2021ringfed} & 89.65\% & 89.53\% & 89.14\% \\
 Fed-Cyclic (ours) & \textcolor{blue}{91.15\%} & \textcolor{blue}{90.89\%} & \textcolor{blue}{90.33\%} \\
 Fed-Star (ours) & \textcolor{red}{91.72\%} & \textcolor{red}{91.17\%} & \textcolor{red}{90.58\%}  \\ 
 \bottomrule
\end{tabular}
}}
\end{center}
\caption{Experimental results show that both the Fed-Star and Fed-Cyclic attains higher accuracy than FedAvg and F1-scores. Here, \textcolor{red}{red} denotes the best value and \textcolor{blue}{blue} denotes the second best value.}
\label{results}
\end{table}


\subsection{Results}
We evaluate all four algorithms (FedAvg~\cite{mcmahan2017communication}, RingFed~\cite{yang2021ringfed}, Fed-Cyclic, Fed-Star) on our dataset. For RingFed, we kept $\gamma$=0.8, as we obtained the best accuracy for RingFed using this value, as shown in Table~\ref{mul_ring}.  

We provide results of both global evaluation and local evaluation. While local test sets are used for local evaluation, their union is used for global evaluation. As we can see in Table~\ref{results}, where we provide the global evaluation results, our two proposed methods, Fed-Cyclic and Fed-Star, perform better than FedAvg and RingFed. Fed-Star performs the best among the four, with an accuracy of 91.72\%. Moreover, our methods converge very fast. Fed-Star requires 50 global rounds on our dataset, as mentioned in Table~\ref{stats3}. Although RingFed also requires the same number of global rounds, its accuracy is lower than Fed-Star.


\begin{table}[ht]
  \begin{center}
    \scalebox{0.95}{{\small{
\begin{tabular}{lcc}
\toprule
Dataset & Model & Test Accuracy \\
\midrule
\multirow{5}{*}{\href{https://www.123rf.com/}{123rf}} & Local Model &85.51\% \\
& FedAvg & 83.79\% \\
& RingFed & 84.47\% \\
& Fed-Cyclic (Ours) & \textcolor{blue}{86.38\%} \\
& Fed-Star (Ours) & \textcolor{red}{88.90\%} \\
\hline
\multirow{5}{*}{\href{https://stock.adobe.com/}{Adobe Stock}} & Local Model & 92.43\% \\
& FedAvg & 91.46\% \\
& RingFed & 92.07\% \\
& Fed-Cyclic (Ours) & \textcolor{blue}{94.52\%} \\
& Fed-Star (Ours) & \textcolor{red}{94.96\%} \\
\hline
\multirow{5}{*}{\href{https://www.alamy.com/}{Alamy}} & Local Model & 86.84\% \\
& FedAvg & 87.32\% \\
& RingFed & 87.98\% \\
& Fed-Cyclic (Ours) & \textcolor{blue}{88.90\%} \\
& Fed-Star (Ours) & \textcolor{red}{90.14\%} \\
\hline
\multirow{5}{*}{\href{https://www.canstockphoto.com/}{CanStockPhotos}} & Local Model & 89.53\% \\
& FedAvg & 89.20\% \\
& RingFed & 89.56\% \\
& Fed-Cyclic (Ours) & \textcolor{blue}{90.40\%} \\
& Fed-Star (Ours) & \textcolor{red}{91.68\%} \\
\hline
\multirow{5}{*}{\href{https://depositphotos.com/}{Depositphotos}} & Local Model & \textcolor{blue}{98.07\%} \\
& FedAvg & 96.95\% \\
& RingFed & 97.76\% \\
& Fed-Cyclic (Ours) & 97.91\% \\
& Fed-Star (Ours) & \textcolor{red}{98.33\%} \\
\hline
\multirow{5}{*}{\href{https://www.gettyimages.in/}{Getty Images}} & Local Model & 89.35\% \\
& FedAvg & 90.13\% \\
& RingFed & 91.06\% \\
& Fed-Cyclic (Ours) &\textcolor{blue} {92.72\%} \\
& Fed-Star (Ours) & \textcolor{red}{93.87\%} \\

\hline
\multirow{5}{*}{\href{https://www.istockphoto.com/}{iStock}} & Local Model & \textcolor{blue}{85.71\%} \\
& FedAvg & 83.72\% \\
& RingFed & 84.68\% \\
& Fed-Cyclic (Ours) & 84.89\% \\
& Fed-Star (Ours) & \textcolor{red}{86.47\%} \\
\hline
\multirow{5}{*}{\href{https://www.shutterstock.com/}{Shutterstock}} & Local Model & 89.45\% \\
& FedAvg & 89.60\% \\
& RingFed & 90.16\% \\
& Fed-Cyclic (Ours) & \textcolor{blue}{91.56\%} \\
& Fed-Star (Ours) & \textcolor{red}{92.11\%} \\
\bottomrule
\end{tabular}
}}}
\end{center}
\caption{Fed-Star outperforms all the local models trained using traditional ML method and baselines and Fed-Cyclic on different sources.}
\label{personal}
\end{table}

In Table~\ref{personal}, where we provide the results of the local evaluation, we compare our methods with competing Federated Learning methods and the baseline of the respective local model (using $E=250$). As can be seen, our Fed-Star algorithm gets the best results on all the clients. Our Fed-Cyclic algorithm comes second in 6 out of 8 cases, losing to the local model on Depositphotos and iStock. As far as FedAvg and RingFed are concerned, they lose to 5 and 4 local clients, respectively. This suggests that FedAvg and RingFed do not help much from a personalization point of view.    


\begin{figure*}[!htb]
  \begin{center}
  \includegraphics[width=0.85\textwidth]{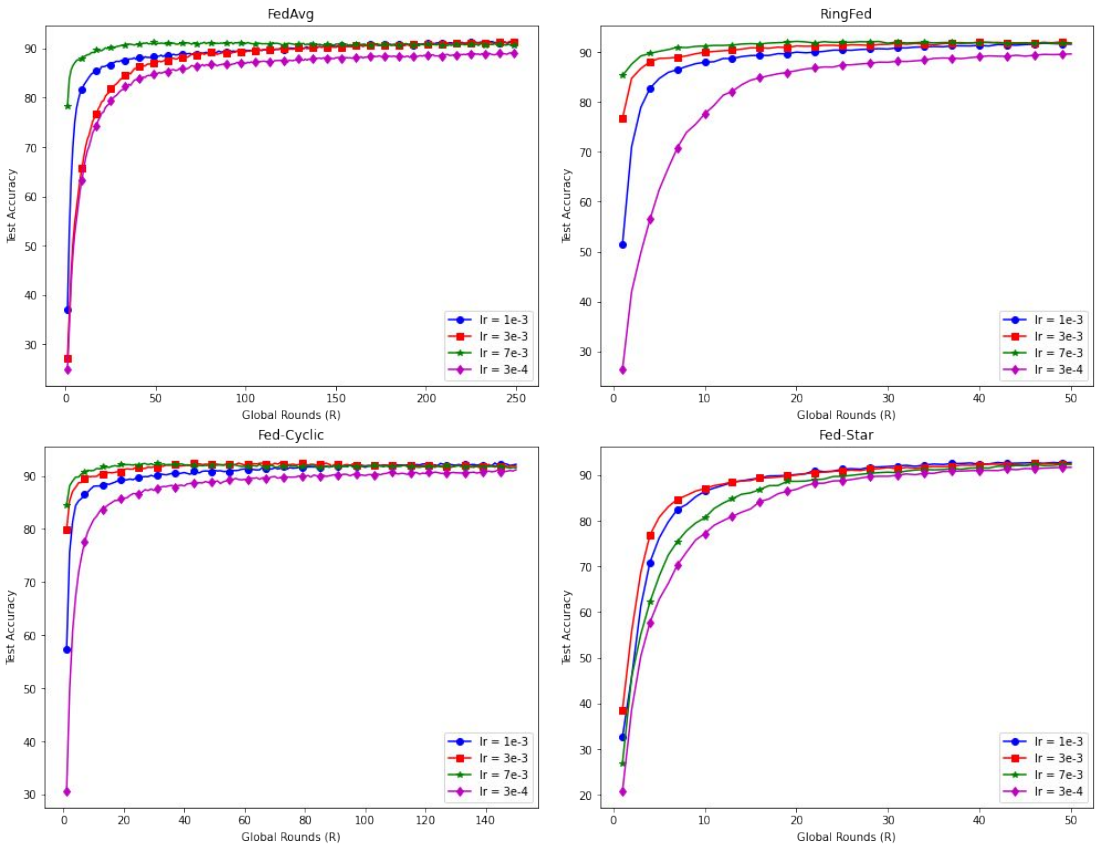}
  \caption{The graph shows how the accuracy of FedAvg, RingFed, Fed-Cyclic and Fed-Star changes with different learning rates (lr).}\label{fig:lr_param}
  \end{center}
  
\end{figure*}

\subsection{Learning Rate ($\eta$) Experiments}

We also performed our experiments while varying the learning rate. For FedAvg, we have observed that accuracy steadily increases with the decrease in the learning rate, and maximum accuracy is obtained for the learning rate of 3e-3 with the value of 91.43\%. For RingFed, we observed that the value of accuracy increased with an increase in learning rate from 1e-3 to 7e-3 (91.81\% to 92.17\%), followed by dropping in accuracy for the learning rate of 3e-4 to 89.65\%. For the Fed-Cyclic algorithm, maximum accuracy is obtained for the learning rate of 3e-3 with an accuracy value of 92.52\% and minimum accuracy of 91.15\% for the learning rate of 3e-4. Fed-Star attains maximum accuracy of 92.77\% for a learning rate of 1e-3. The accuracy drops with an increase in the learning rate, falling to the value of 91.72\% for the learning rate of 3e-4. The detailed results are captured in Figure~\ref{fig:lr_param} and Table~\ref{tab:lr_rates}.




\begin{table}[!htb]
  \begin{center}
    {
\begin{tabular}{ccccc}
\toprule 
 Learning & FedAvg & RingFed   & Fed-Cyclic & Fed-Star \\
  Rate ($\eta$) &~\cite{mcmahan2017communication}  &~\cite{yang2021ringfed} & (Ours) & (Ours) \\
 \midrule
 1e-3 & 91.39\% & 91.81\% & \textcolor{blue}{92.42\%} & \textcolor{red}{92.77\%} \\ 
 3e-3 & 91.43\% & 92.09\% & \textcolor{blue}{92.52\%} & \textcolor{red}{92.68\%} \\
 7e-3 & 91.33\% & 92.17\% & \textcolor{blue}{92.35\%} & \textcolor{red}{92.53\%} \\
 3e-4 & 89.11\% & 89.65\% & \textcolor{blue}{91.15\%} & \textcolor{red}{91.72\%} \\ 
 \bottomrule
\end{tabular}
}
\end{center}
\caption{The table shows the accuracy value after convergence attained by FedAvg, RingFed, Fed-Cyclic and Fed-Star for different values of learning rates.}\label{tab:lr_rates}
\end{table}
\section*{Conclusion}
We proposed two Federated Learning algorithms, Fed-Cyclic and Fed-Star, along with a new federated image classification dataset collected from 8 commercial image sources, making the setup much closer to a real-world scenario than other image classification setups where an existing dataset itself is artificially divided. Our algorithms have better convergence and better accuracy than FedAvg and RingFed algorithms. Also, they perform much better from the personalization point of view, making them very relevant for meaningful collaboration amongst clients having statistical heterogeneity (domain shift).  

{\small
\bibliographystyle{ieee_fullname}
\bibliography{main}
}
\end{document}